\pdfoutput=1

\documentclass[11pt]{article}

\usepackage{acl}

\usepackage{times}
\usepackage{latexsym}

\usepackage[T1]{fontenc}

\usepackage[utf8]{inputenc}

\usepackage{times}
\usepackage{latexsym}
\usepackage{booktabs}
\usepackage{amsmath}
\usepackage{float}
\usepackage{xcolor}
\usepackage{adjustbox}
\usepackage{longtable}
\usepackage{graphicx} 
\usepackage{placeins} 
\usepackage{microtype} 
\usepackage{enumitem} 
\usepackage{amsmath, bm} 
\usepackage{comment} 
\usepackage{amsmath, bm} 
\DeclareMathOperator*{\argmax}{argmax}

\usepackage{microtype}

\title{Consistency and Coherence from Points of Contextual Similarity}

\author{Oleg Vasilyev, John Bohannon \\
  Primer Technologies Inc. \\
  San Francisco, California \\
  \texttt{{oleg,john}@primer.ai}\\}

\begin{document}
\maketitle
\begin{abstract}
Factual consistency is one of important summary evaluation dimensions, especially as summary generation becomes more fluent and coherent. The ESTIME measure, recently proposed specifically for factual consistency, achieves high correlations with human expert scores both for consistency and fluency, while in principle being restricted to evaluating such text-summary pairs that have high dictionary overlap. This is not a problem for current styles of summarization, but it may become an obstacle for future summarization systems, or for evaluating arbitrary claims against the text. In this work we generalize the method, and make a variant of the measure applicable to any text-summary pairs. As ESTIME uses points of contextual similarity, it provides insights into usefulness of information taken from different BERT layers. We observe that useful information exists in almost all of the layers except the several lowest ones. For consistency and fluency - qualities focused on local text details - the most useful layers are close to the top (but not at the top); for coherence and relevance we found a more complicated and interesting picture.
\end{abstract}

\section{Introduction}
A summary is assessed by evaluating its qualities, which can be defined in different ways \cite{Fan:2018:Robust, Xenouleas:2019:SUMQE, Kryscinski:2020:EvaluatingFactual,  Vasilyev:2020:Sensitivity, Fabbri:2021:SummEval}. The commonly considered qualities are of two classes: summary-focused and summarization-focused. The summary-focused qualities are supposed to reflect the language of the summary itself, without any relation to the summarization. For example, grammar, fluency, structure and coherence of a summary should not require consideration of the text from which the summary is produced: grading a summary in this respect is no different from evaluation of any other text, generated or not. The summarization-focused qualities are supposed to reflect the summarization, and require both the summary and the text considered together, - for example, relevancy, informativeness, factual consistency. 

The automated evaluation measures normally use both the summary and the text \cite{Louis:2009:JShannon, Scialom:2019:SummaQA, Gao:2020:SUPERT, Vasilyev:2020:BLANC, Vasilyev:2020:ESTIME, Scialom:2021:QuestEval}; other measures use the summary and the 'reference summaries' - the summaries human-written specifically for the text \cite{Papineni:2002:BLEU, Lin:2004:ROUGE, Zhang:2020:BERTScore}. Correlation of evaluation measures with human scores for all qualities is widely accepted as a criterion for judging about the evaluation measures \cite{Fabbri:2021:SummEval}, with a few caveats.

Since the summary qualities (e.g. relevance, consistency, coherence and fluency) are all different, improving a measure eventually cannot be expected to keep improving the correlations with all the qualities. 
It would be natural to have measures intentionally focused on certain qualities. 
If a measure correlates reasonably well (by current standards) with all the qualities, this may reflect the simple reality that better generation systems produce summaries with better qualities overall, and that easier texts allow to produce summaries with better qualities overall. There is also a possibility of an implicit bias even in expert scores \cite{Vasilyev:2021:NoHuman}.

Factual consistency is arguably the most objective quality of the summary, and there were consistent efforts to improve its evaluation \cite{Falke:2019:Ranking, Kryscinski:2020:EvaluatingFactual, Wang:2020:Asking, Maynez:2020:Faithfulness, Scialom:2021:QuestEval, Gabriel:2021:GoFigure}. Recently introduced measure ESTIME is focused on factual consistency, and achieves superior scores both for consistency and fluency \cite{Vasilyev:2020:ESTIME}\footnote{https://github.com/PrimerAI/blanc/tree/master/estime}. The measure is simple, interpretable and easily reproducible: It is not tuned on some specific human-annotated dataset but simply using a well known pretrained language model, and almost no parameters to set or chose. It can use any other pretrained language model, including multilingual BERT, thus being ready for other languages. 
However, ESTIME can be used only in situations where the summary and the text have high dictionary overlap. This is not a problem for evaluating current generation systems (which rarely introduce words not from the text), but this makes more problematic a progress toward estimation of factual consistency of an arbitrary 'claim' (not necessarily a summary) with respect to a text.

Our contribution\footnote{The code for all the introduced measures will be added to https://github.com/PrimerAI/blanc/tree/master/estime}:
\begin{enumerate}[topsep=0pt,itemsep=-1ex,partopsep=1ex,parsep=1ex]
    \item We generalize ESTIME, and present its variant applicable to more scenarios.
    \item We get insights on usefulness of embeddings from different BERT layers for evaluating summary qualities.
    \item We provide an alternative measure for evaluating coherence. Curiously, its correlations with human scores are better when using the lower-middle part of large BERT.
\end{enumerate}

\section{Evaluation at points of similarity}
ESTIME as defined in \cite{Vasilyev:2020:ESTIME} can be expressed by a count of all the instances of mismatched tokens at the points of similarity between the summary and the text:

\begin{equation} \label{eq:n_alarms2}
\sum_{i: t_i \in T}(1 - \delta_{t_it'_{\argmax_{\alpha}(e_ie'_\alpha)}})
\end{equation}

The summary is represented by a sequence of tokens $t_i$, each having embedding $e_i$; the text $T$ is a sequence of tokens $t'_\alpha$, having embeddings $e'_\alpha$. The tokens are the words or word parts as defined by a pretrained language model that is used to obtain the embeddings. For each summary token $t_i$, located at a point $i$, there is a point of similarity $\beta = \argmax_{\alpha}(e_ie'_\alpha)$, and the similarity is defined by the product of the embeddings $(e_ie'_\alpha)$. If the tokens $t_i$ and $t'_\beta$ do not coincide, it adds up to the total count of 'alarms'. The summation in Equation \ref{eq:n_alarms2} is over all the summary tokens $t_i$ that exist in the text $T$. 

The embeddings $e$ are the contextual embeddings, obtained from processing the \emph{masked} tokens. Loosely speaking, ESTIME imitates what a human does to find factual inconsistencies: for each point in a summary, find or recall the point of similar context in the text, and verify that the summary responds to this context in the same way as the text does. In this sense, any 'fact' is simplistically represented by a duple of context and response to the context, which is a token. ESTIME verifies that all such summary facts are in agreement with their counterparts in the text. 

We replace the response: instead of the tokens $t_i$, we consider raw (input) token embeddings $\phi_i$. We define 'ESTIME-soft' consistency score as:

\begin{equation} \label{eq:estime_soft}
\frac{1}{N} \sum_{i=1}^Ncos(\phi_i, \phi'_{\argmax_{\alpha}(e_ie'_\alpha)})
\end{equation}

The agreement is now measured not between the tokens, but between the raw token embeddings at the points of similarity. The summation is over all the summary tokens, and the result is the averaged agreement: $N$ is the length of the summary in terms of the tokens. 

The context similarity is measured using dot-product of the contextualized embeddings $e$ of masked tokens, and the embeddings are to be taken from one of the layers of the model, the same way as it is done in ESTIME. The length of the embeddings is influenced by the context, and this influence is not lost by dot-product. But the agreement between the 'responses' is measured by the cosine of the angle between the raw token embeddings $\phi$: the context is not needed here.

\section{Correlations with expert scores}\label{sec:Correlations}

\begin{table*}[th]
\centering
\begin{tabular}{@{}lllllllll@{}}
\toprule
\textbf{measure} & \multicolumn{2}{c}{consistency} & \multicolumn{2}{c}{relevance} &
\multicolumn{2}{c}{coherence} & \multicolumn{2}{c}{fluency}\\
\midrule
{} & \bm{$\rho$} & \bm{$\tau$}
 & \bm{$\rho$} & \bm{$\tau$}
  & \bm{$\rho$} & \bm{$\tau$}
   & \bm{$\rho$} & \bm{$\tau$}\\
BLANC & 0.19 & 0.10 & 0.28 & 0.20 & 0.22 & 0.16 & 0.13 & 0.07\\
ESTIME & 0.39 & 0.19 & 0.15 & 0.11 & 0.27 & 0.19 & \textbf{0.38} & \textbf{0.22}\\
ESTIME-soft & \textbf{0.40} & \textbf{0.20} & 0.25 & 0.18 & 0.28 & 0.20 & 0.36 & 0.21\\
Jensen-Shannon & 0.18 & 0.09 & 0.39 & 0.28 & 0.29 & 0.21 & 0.11 & 0.06\\
SummaQA-F & 0.17 & 0.08 & 0.14 & 0.10 & 0.08 & 0.06 & 0.12 & 0.07\\
SummaQA-P & 0.19 & 0.09 & 0.17 & 0.12 & 0.10 & 0.08 & 0.12 & 0.07\\
SUPERT & 0.28 & 0.14 & 0.26 & 0.19 & 0.20 & 0.15 & 0.17 & 0.10\\
\hline
BERTScore-F & 0.10 & 0.05 & 0.38 & 0.28 & \textbf{0.39} & \textbf{0.28} & 0.13 & 0.07\\
BERTScore-P & 0.05 & 0.03 & 0.29 & 0.21 & 0.34 & 0.25 & 0.11 & 0.06\\ 
BERTScore-R & 0.15 & 0.08 & \textbf{0.41} & \textbf{0.30} & 0.34 & 0.25 & 0.11 & 0.06\\
BLEU & 0.09 & 0.04 & 0.23 & 0.17 & 0.19 & 0.14 & 0.12 & 0.07\\
ROUGE-L & 0.12 & 0.06 & 0.23 & 0.16 & 0.16 & 0.11 & 0.08 & 0.04\\
ROUGE-1 & 0.13 & 0.07 & 0.28 & 0.20 & 0.17 & 0.12 & 0.07 & 0.04\\
ROUGE-2 & 0.12 & 0.06 & 0.23 & 0.16 & 0.14 & 0.10 & 0.06 & 0.04\\
ROUGE-3 & 0.15 & 0.07 & 0.23 & 0.17 & 0.15 & 0.11 & 0.06 & 0.04\\
\bottomrule
\end{tabular}
\caption{Summary level correlations $\rho$ (Spearman) and $\tau$ (Kendall Tau-c) of quality estimators with human experts scores. Truly automated evaluation measures are in the top rows; the measures under the separation line need human-written reference summaries. In each column the highest correlation is bold-typed.}
\label{tab:summary-level}
\end{table*}

We compare ESTIME-soft with other measures by using SummEval dataset \footnote{https://github.com/Yale-LILY/SummEval} \cite{Fabbri:2021:SummEval}. The human-annotated part of the dataset consists of 100 texts, each text is accompanies by 17 summaries generated by 17 systems. All 1700 text-summary pairs are annotated (on scale 1 to 5) by 3 experts for 4 qualities: consistency, relevance, coherence and fluency. 
Each text is also accompanied by 11 human-written reference summaries, for the measures that need them.

Correlations of a few measures with average expert scores (each score is an average of the three individual expert scores) are shown in Table \ref{tab:summary-level}. The results for most measures and qualities are a bit different from \cite{Vasilyev:2020:ESTIME}, where the original version of SummEval was used, with 16 (not 17) generation systems. BLANC is taken by its default version\footnote{https://github.com/PrimerAI/blanc}; ESTIME is taken with embeddings taken from layer 21. ESTIME-soft is obtained using the same model 'bert-large-uncased-whole-word-masking' and the same layer 21 for the points of contextual similarity, while using 'bert-base-uncased' for the raw embeddings. More details on calculating the correlations of Table \ref{tab:summary-level} are in Appendix \ref{app:Correlations}.

ESTIME-soft has in Table \ref{tab:summary-level} approximately the same correlations with expert scores as ESTIME. The distinct improvement for relevance is not important by itself: a summary-to-text verification measure should not be applied to judge relevance. But together with small improvements in consistency and coherence this suggests that context is used better in ESTIME-soft than in ESTIME, in expense of correlations with fluency. Of course ESTIME and ESTIME-soft can be calculated together, because the main processing cost is in obtaining the masked embeddings, which are used then for both of the measures.

\section{Embeddings from different layers}
\subsection{Layers for summary and text}\label{sec:Layers-summ-text}

\begin{figure*}[]
\centering
\includegraphics[width=\linewidth]{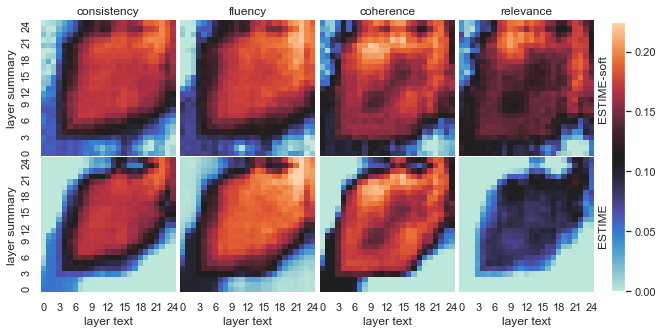}
\caption{Kendall Tau-c correlation between ESTIME and SummEval expert scores. Upper row is for ESTIME-soft, lower row is for ESTIME. Each column is for one of the qualities. Points of similarity are found by using embeddings (of \emph{masked} tokens) from BERT layer of axis X for text, and layer of axis Y for summary.}
\label{Fig_Kendall_AllQualities}
\end{figure*}

We can expect that our measure given by Eq.\ref{eq:estime_soft} is not only applicable in more situations, but also more robust. In Figure \ref{Fig_Kendall_AllQualities} we show how the correlations with expert scores depend on the BERT layers from which the embeddings $e$ are taken. We allow embeddings for the summary and the text to be taken from different layers. Our motivation for this is that (1) the summary and the text have different information density, and (2) we may get more insight at how robust the measures are. Figure \ref{Fig_Kendall_AllQualities} shows Kendall Tau-c correlations. Spearman correlations manifest similar patterns, see Appendix \ref{app:Spearman}.

Indeed, we observe that for the consistency ESTIME-soft is more tolerant to the choice of the layers, including the higher and more flat peak around the layers 21 (compared to ESTIME). For all qualities ESTIME-soft is wider across the layers. Compared to consistency, the fluency peak at high levels is more pronounced, this might be because the fluency is closer to the task of tokens prediction, which is the actual goal of the BERT training. 

Coherence and relevance are the qualities that, on opposite, require long-range context for making a judgement. It is interesting therefore that ESTIME-soft (and ESTIME) correlates the best with both of these qualities away from the diagonals in Figure \ref{Fig_Kendall_AllQualities}, when the embeddings for the summary tokens and for the text tokens come from different layers. The shape of these heatmaps suggests that something is happening at the middle of the transformer, and that embeddings from different layers may have complementary information, beneficial for finding the points of contextual similarity, and ultimately to the correlations. 

The usefulness of ESTIME-soft Eq.\ref{eq:estime_soft} depends on two factors: (1) successful finding of the points of contextual similarity $(e_ie'_{\alpha})$, where we now allow $e_i$ and $e_{\alpha}'$ to be picked up from different layers, and (2) a reliable relation between the context $e_i$ and the 'response' $\phi_i$ (likewise between $e_{\beta}'$ and $\phi_{\beta}'$). We speculate that satisfying these two factors using a single transformer layer is easier for a 'local context' quality - consistency or fluency, - but it is more difficult for coherence or relevance, for which a longer range context is important.

\subsection{Changes across the layers}\label{sec:layers-changes}

The coherence and relevance correlations with ESTIME in Figure \ref{Fig_Kendall_AllQualities} suggests that there may be a qualitative difference between the context-relevant information in the lower half and the upper half of the layers. Transformer layers were probed in \cite{Zhang:2020:BERTScore, Vasilyev:2020:ESTIME} by their usefulness for estimating qualities of a summary; the layers were also inspected by other kinds of 'probes' in \cite{Jawahar:2019:LanguageStructure, Voita2019Evolution, Liu:2019:Linguistic, Kashyap:2019:DomainRobustness, Aken:2019:QuestionsLayers, Wallat:2020:bertnesia, Cao2020Decisions, Timkey:2021:RogueDimensions,  Turton:2021:BinderEmbeddings,   Geva:2021:TransformerLayers, Yun2021Transformer, Fomicheva2021Translation}. 

\begin{figure}[h]
\centering
\includegraphics[width=\linewidth]{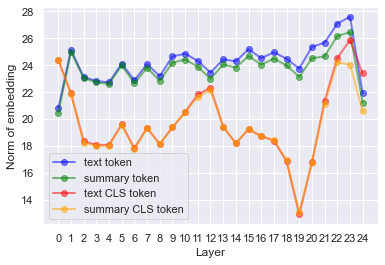}
\caption{Norm of embedding $e^k_i$ of a masked token $i$, for each layer $k$. The token is either CLS or the usual text token; the token is also either taken from a summary or from a text. Results are averaged over all tokens of each summary or text, and then the results are averaged over all the summaries or texts (of SummEval dataset). Model: 'bert-large-uncased-whole-word-masking'.}
\label{Fig_Layers_norm}
\end{figure}

Since the points of contextual similarity are defined in Eqs.\ref{eq:n_alarms2} and \ref{eq:estime_soft} by dot-product of the embeddings of the masked tokens, we are curious about the evolution of such embeddings through the layers, regardless of the summarization task. The embedding norm is one interesting characteristic, as being solely responsible for the peak around the layer 21 for the correlations with consistency \cite{Vasilyev:2020:ESTIME}. In Figure \ref{Fig_Layers_norm} we present the norm of masked token embeddings across the layers, using the same annotated part of SummEval dataset: 100 texts and 1700 generated summaries.
Notice that we are using these texts and summaries simply as separate texts, and simply considering the embeddings of all masked tokens of the text.

We observe that there is only minor difference between considering the tokens of the text and the tokens of the summary. The embeddings of the text tokens have slightly higher norm than embeddings of the summary tokens, probably boosted by having more related context for each token.
More importantly, the CLS token, having more long-range semantic information, gets a big peak of the norm exactly in the middle of the transformer, and a big dip and peak at higher layers.

\begin{figure}[h]
\centering
\includegraphics[width=\linewidth]{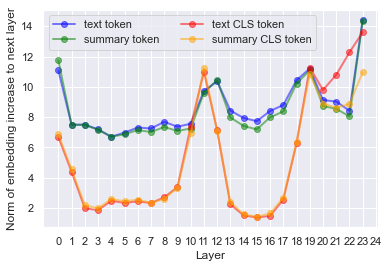}
\caption{Norm of the change of embedding. The data and the notations are the same as in Figure \ref{Fig_Layers_norm}, except that here for layer $k$ we show $|e^{k+1}_i - e^{k}_i|$: the norm of the difference of embeddings of the same (masked) token $i$ from the neighbor layers.}
\label{Fig_Layers_norm_tonext}
\end{figure}

In Figure \ref{Fig_Layers_norm_tonext} we consider how much does an embedding changes from one layer to the next, on average. 
The figure shows the norm of the increment of a masked token embedding from one layer to the next. The change exactly in the middle of the transformer shows up even stronger.

\begin{figure}[h]
\centering
\includegraphics[width=\linewidth]{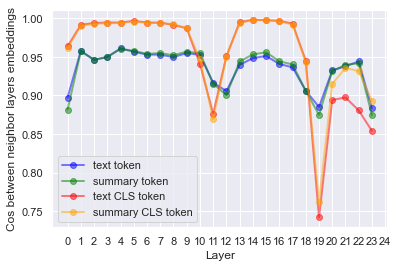}
\caption{Cosign between embeddings from neighbor layers. The data and the notations are the same as in Figure \ref{Fig_Layers_norm}, except that here for layer $k$ we show $cos(e^{k+1}_i, e^{k}_i)$: the cosign of the angle between embeddings of the same (masked) token $i$ from the neighbor layers.}
\label{Fig_Layers_cos_tonext}
\end{figure}

A cosine between the embeddings of the same token in two neighbor layers is shown in Figure \ref{Fig_Layers_cos_tonext}. 
There are almost no changes in the CLS token across the layers, except the strong change in the middle layers 10-13, and then at high layers starting from layer 18. The common text tokens keep going through substantial transformation, remarkably with a stable pace - the cosign between the embeddings of neighbor layers is about 0.95, - and with much larger changes at the few special layers at which the CLS embedding also changes.

The evolution of the masked token embedding across the layers reflected in Figures \ref{Fig_Layers_norm}, \ref{Fig_Layers_norm_tonext}, \ref{Fig_Layers_cos_tonext} is not related to the summarization: this is how the embeddings change for a text or a summary (see also Appendix \ref{app:Layers_embeddings}). Looking in retrospect on the coherence in Figure \ref{Fig_Kendall_AllQualities}, it seems natural to split the layers-by-layers square into four quadrants.

\section{Coherence and points of similarity}\label{sec:Coherence}
\subsection{Coherence}\label{sec:Coherence_measure}
As we have seen in Figure \ref{Fig_Kendall_AllQualities}, the coherence and relevance would get higher correlations with expert scores when the embeddings are picked up from different layers. The layers 18-21 for summary and around 8-11 for text would give a better correlations for coherence than in Table \ref{tab:summary-level}. We may speculate that the reason for this is that coherence and relevance are not as 'local' as consistency and fluency, and need a more varied information for a judgement. When forced to use the points of similarity between the summary and the text, they do better with different layers.

\begin{figure}[h]
\centering
\includegraphics[width=\linewidth]{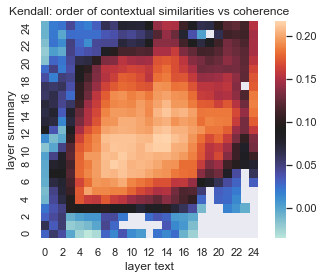}
\caption{Kendall Tau-c correlations between the order of summary-text similarity points (Eq.\ref{eq:estime_order}) and expert coherence scores. Points of similarity are found by embeddings from BERT layer of axis X for text, and layer of axis Y for summary.}
\label{Fig_Order_Coherence_Kendall}
\end{figure}

In principle there should be a way to estimate the coherence and fluency of a summary without using the text, but having a text as a helpful pattern makes it easier. For better understanding of how the text helps, we can consider estimation of coherence by using the order of the points of similarity. Let us consider a correlation between the sequence of summary token locations $i = 0, 1, 2, ..., N$ and the sequence of the corresponding similarity locations $\argmax_{\alpha}(e_ie'_\alpha)$ of the tokens in the text:
\begin{equation} \label{eq:estime_order}
\tau\left((i)_{i=1}^N, (\argmax_{\alpha}(e_ie'_\alpha))_{i=1}^N\right)
\end{equation}
Equation \ref{eq:estime_order}, with $\tau$ being Kendall Tau-c correlation, estimates how well the summary keeps the order of the points of similar context. Figure \ref{Fig_Order_Coherence_Kendall} shows correlations of this measure with expert coherence scores. 
This figure is very different from the coherence part of Figure \ref{Fig_Kendall_AllQualities}. For once, the diagonal - using the same layers for the summary and the text - is fine again. The plateau of high correlations is wide. 
Using the same layer for the summary and the text, we get correlation with coherence 0.20 or 0.21 in the layers from 7 to 15, which are quite far from the top layers. These layers, however, would be in the top part of a base BERT.

The coherence estimate by Eq.\ref{eq:estime_order} involves only the points of similar context, and (unlike Eqs.\ref{eq:n_alarms2} and \ref{eq:estime_soft}) does not deal with comparing the 'responses'. Such context happens to be better represented for coherence by the layers in the lower-middle part of the large BERT.

\subsection{Local order of similarities}\label{sec:Coherence_local}
For a better insight into how the consistency is a more 'local' quality than coherence, we will modify Kendall Tau correlation between two sequences by introducing a restriction on the distance between the elements of the first sequence. In applying such a modified Kendall correlation in Eq.\ref{eq:estime_order}, we would consider only such pairs of summary tokens locations $i$ that are not further from each other than an (integer) interval $d$:
\begin{equation} \label{eq:estime_order_local}
\tau_d\left((i)_{i=1}^N, (\argmax_{\alpha}(e_ie'_\alpha))_{i=1}^N\right)
\end{equation}
In Eq.\ref{eq:estime_order_local} we use the modified 'local' Kendall Tau correlation $\tau_d$. We define $\tau_d$ as a 
simple Kendall Tau correlation between two sequences $X$ and $Y$, but with a restriction on the distance between the elements $X$. As usual, $\tau_d = \frac{n_c - n_d}{n_{t}}$, where $n_c$ is the number of concordant pairs, $n_d$ is the number of discordant pairs, and $n_t$ is the total number of pairs, but the counts include only the pairs $\{(X_i,Y_i),(X_j,Y_j)\}$ for which $|X_i - X_j|<=d$. 
The concordant pair is a pair satisfying the condition $(X_i>X_j \; \textrm{and} \; Y_i>Y_j) \; \textrm{or} \; (X_i<X_j \; \textrm{and} \; Y_i<Y_j)$, otherwise the pair is discordant.
Note that in our case (Eq.\ref{eq:estime_order_local}) the distance is the same no matter whether it is defined between the $X$ elements or between the $X$ ranks.

\begin{figure}[h]
\centering
\includegraphics[width=\linewidth]{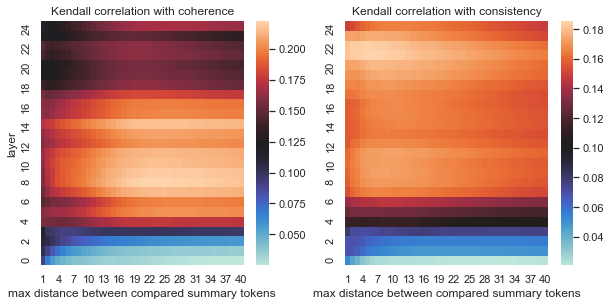}
\caption{Kendall Tau-c correlations between the 'local order' measure Eq.\ref{eq:estime_order_local} and SummEval expert scores of coherence and consistency. The correlations depend on the measure distance restriction $d$ (axis X) and the layer from which the embeddings are taken (axis Y).}
\label{Fig_KendallRange}
\end{figure}

Figure \ref{Fig_KendallRange} shows the correlations of the measure given by Eq.\ref{eq:estime_order_local} with SummEval expert scores of coherence and consistency. 
We can make several simple conclusions from this figure and its underlying data:
\begin{enumerate}[topsep=0pt,itemsep=-1ex,partopsep=1ex,parsep=1ex]
    \item Consistency is a more 'local' quality than coherence: keeping order on shorter distances provides better correlations. For consistency the layer 21 provides the best correlations - exceeding 0.18 - at the distances from 1 to 15. For coherence the layer 8 provides the best correlations - exceeding 0.22 - at the distances from 19 to 33.
    \item There is a wide range of layers that are helpful both for coherence and for consistency. The best layers are at the lower half of the BERT for the coherence, and close to the top for the consistency.
    \item The verification of facts (responses) by Eq.\ref{eq:estime_soft} or \ref{eq:n_alarms2} is not important (and maybe not needed) for coherence: estimation of order of similarities by Eq.\ref{eq:estime_order_local} or \ref{eq:estime_order} is sufficient. Naturally, for consistency the measures by Eq.\ref{eq:estime_soft} or \ref{eq:n_alarms2}, based on a (simplistic) verification of facts, provide better correlations.
\end{enumerate}

\section{Conclusion}\label{sec:Conclusion}
In this paper we presented ESTIME-soft, Eq.\ref{eq:estime_soft}, - a variant of ESTIME applicable to more scenarios of estimating factual consistency and less sensitive to the choice of the BERT layers for the embeddings. We considered estimation of consistency as verification that the summary and the text respond the same way in all points of similar context. Each 'fact' is represented, accordingly, as a context and a response to the context.

The context similarity for consistency and fluency has a better correlations with human scores when it uses embeddings from the BERT layers close to the top (but not at the top). For coherence the embeddings are more helpful when taken from different layers for the summary and for the text. More robust estimation of coherence is obtained by inspecting the order of the points of contextual similarity (Esq.\ref{eq:estime_order}, \ref{eq:estime_order_local}). Embeddings from lower-middle part of BERT turned out to be the most helpful for this estimation.

A systematic comparison of evaluation measures would have to include much more measures (e.g. \cite{Fabbri:2021:QAFactEval, Fabbri:2021:SummEval, Gabriel:2021:GoFigure}). 
From our limited observations we suggest that there must be much better ways to evaluate the summary qualities, and achieve higher correlations with human scores. 
If currently superior correlations with human scores can be achieved by a simple measure that is based on an oversimplified representation of consistency or coherence and does not need specially tuned models or specially prepared data, then we probably should raise expectations.


\bibliography{anthology,custom}
\bibliographystyle{acl_natbib}

\appendix

\section{Calculation of correlations of selected measures with expert scores}
\label{app:Correlations}
Table \ref{tab:summary-level} in Section \ref{sec:Correlations} shows two types of rank correlations: Kendall Tau-c correlations and Spearman correlations. We do not consider non-rank correlations like Pearson because such presentations could be misleading: (1) there is a subjectivity in the intervals between the '1-2-3-4-5' grades in human scoring, and (2) different automated measures may be differently dilated or contracted at different intervals of their ranges - which is irrelevant for their judgements in comparing the summaries.  

BLANC \cite{Vasilyev:2020:BLANC} is calculated as default BLANC-help\footnote{https://github.com/PrimerAI/blanc}. ESTIME and Jensen-Shannon \cite{Louis:2009:JShannon} values are negated. 
SummaQA \cite{Scialom:2019:SummaQA} is represented by SummaQA-P (prob) and SummaQA-F1 (F1 score)\footnote{https://github.com/recitalAI/summa-qa}. 
SUPERT \cite{Gao:2020:SUPERT} is calculated as single-doc with 20 reference sentences 'top20'\footnote{https://github.com/yg211/acl20-ref-free-eval}. 
BLEU \cite{Papineni:2002:BLEU} is calculated with NLTK. BERTScore \cite{Zhang:2020:BERTScore} (by default \footnote{https://github.com/Tiiiger/bert\_score} using roberta-large) is represented by F1, precision (P) and recall (R).
ROUGE \cite{Lin:2004:ROUGE} is calculated with rouge-score package\footnote{https://github.com/google-research/google-research/tree/master/rouge}.

\section{Alignment of embeddings across the layers}
\label{app:Layers_embeddings}
In Section \ref{sec:layers-changes} we have seen that embeddings of the \emph{masked} tokens of texts (regardless of summarization) undergo sharp changes between a few layers, - for example the change of a cosine between neighbor layer embeddings of the same token in Figure \ref{Fig_Layers_cos_tonext}. 
If a cosine of the angle between embeddings within the same layer would make a good characterization of the layer, we would expect to observe different distributions of the cosine at the layers before and after the sharp change in Figure \ref{Fig_Layers_cos_tonext}. 

\begin{figure}[h]
\centering
\includegraphics[width=\linewidth]{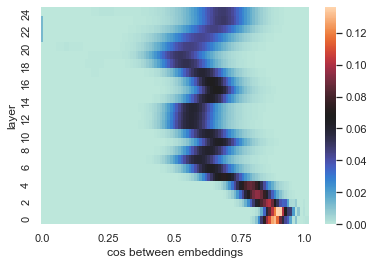}
\caption{Distribution of a cosine between embeddings of two (masked) tokens, taken over all pairs of tokens in each layer, and normalized for each layer. The bar for cosine histogram at each layer is 0.01. This heatmap is obtained from all the SummEval texts and summaries that were used in this paper. There is no any noticeable difference if only texts or only summaries were used here.}
\label{Fig_layers_cos_distribution}
\end{figure}

However, the distribution, shown in Figure \ref{Fig_layers_cos_distribution}, is more involved.
Going from the input up, once the embeddings acquire the context and become on average less aligned with each other, they pass through three full periods of becoming less aligned and then more aligned again. Only the upper two dips in alignment correspond to the sharp changes in Figure \ref{Fig_Layers_cos_tonext}.


\begin{figure*}[]
\includegraphics[width=\linewidth]{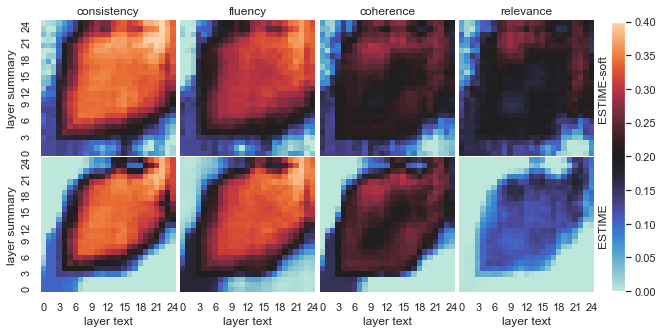}
\caption{Spearman correlation between ESTIME and SummEval expert scores. Upper row is for ESTIME-soft, lower row is for ESTIME. Each column is for one of the qualities. Points of similarity are found by using embeddings (of \emph{masked} tokens) from BERT layer of axis X for text, and layer of axis Y for summary.}
\label{Fig_Spearman_AllQualities}
\end{figure*}

\section{Spearman correlations: dependency on layers}
\label{app:Spearman}
In section \ref{sec:Layers-summ-text} we supported our discussion with figures showing Kendall Tau-c correlations.
Figure \ref{Fig_Spearman_AllQualities} shows the corresponding Spearman correlations. A notable difference between the figures \ref{Fig_Spearman_AllQualities} and \ref{Fig_Kendall_AllQualities} is that in the Spearman version the coherence and the relevance lost more of their level of correlations as compared to the consistency and fluency. We interpret this as a consequence of that Kendall's $\tau$ (based on all pairs of observations) is more helpful than Spearman's $\rho$ in accounting for overall long-range correlations. Thus, Kendall's $\tau$ delivers more for coherence and relevance (as compared to the 'local' qualities - consistency and fluency).

\end{document}